\documentclass[conference]{IEEEtran}
\IEEEoverridecommandlockouts
\usepackage{cite}
\usepackage{amsmath,amssymb,amsfonts}
\usepackage{algorithmic}
\usepackage{graphicx}
\usepackage{amsmath}
\usepackage{amssymb}
\usepackage[utf8]{inputenc}
\usepackage{float}
\usepackage{caption}
\usepackage{subcaption}
\usepackage{tabularx}
\usepackage{booktabs} 
\usepackage{textcomp}
\usepackage{xcolor}
\def\BibTeX{{\rm B\kern-.05em{\sc i\kern-.025em b}\kern-.08em
    T\kern-.1667em\lower.7ex\hbox{E}\kern-.125emX}}
  
\def \paperPath {}
    
\begin{document}

\title{Prediction Confidence from Neighbors\\
}

\author{\IEEEauthorblockN{1\textsuperscript{st} Mark P. Philipsen}
\IEEEauthorblockA{\textit{Danish Meat Research Institute} \\
\textit{Danish Technological Institute}\\
Taastrup, Denmark \\
mpph@create.aau.dk}
\and
\IEEEauthorblockN{2\textsuperscript{nd} Thomas B. Moeslund}
\IEEEauthorblockA{\textit{Visual Analysis of People Group} \\
\textit{Aalborg University}\\
Aalborg, Denmark \\
tbm@create.aau.dk}
}

\maketitle

\begin{abstract}
The inability of Machine Learning (ML) models to successfully extrapolate correct predictions from out-of-distribution (OoD) samples is a major hindrance to the application of ML in critical applications. Until the generalization ability of ML methods is improved it is necessary to keep humans in the loop. The need for human supervision can only be reduced if it is possible to determining a level of confidence in predictions, which can be used to either ask for human assistance or to abstain from making predictions.
We show that feature space distance is a meaningful measure that can provide confidence in predictions. The distance between unseen samples and nearby training samples proves to be correlated to the prediction error of unseen samples. Depending on the acceptable degree of error, predictions can either be trusted or rejected based on the distance to training samples. 
This enables earlier and safer deployment of models in critical applications and is vital for deploying models under ever-changing conditions.
\end{abstract}

\begin{IEEEkeywords}
Deep Learning, Confidence, Abstention, Out-of-distribution detection
\end{IEEEkeywords}

\section{Introduction}
Deploying Machine Learning (ML) methods, most noticeably Deep Neural Networks (DNNs), in a complex and changing world reveals some fundamental shortcomings with these methods. Focusing on DNNs, the shortcomings include failures in classifying known objects when presented with novel poses~\cite{DBLP:journals/corr/abs-1811-11553} and a lack of robustness to minor perturbations in the input, where small changes, imperceptible to humans, may result in incorrect predictions~\cite{szegedy2013intriguing}. Both problems can be mitigated with training data that densely covers the experience space. Ensuring this often proves challenging, especially in dynamic environments.
Such environments are called non-stationary and are likely to result in what is known as concept drift, which occurs when data distributions change such that existing models become outdated and must be retrained. In less severe cases this is called virtual concept drift and will require tuning of representations or supplemental learning~\cite{elwell2011incremental}. Spam detection~\cite{nosrati2011dynamic}, price prediction~\cite{harries1995detecting}, and factory automation~\cite{lin2019concept,pechenizkiy2010online}, are examples of applications that encounter concept drift and where it is necessary to retrain models from time to time.

For the vast majority of ML applications it is assumed that the world is predictable and static. It may seem that way most of the time, but the reality is that things can suddenly change in surprising ways. Data sets used during system development and used in benchmarks are usually captured with a low variety in hardware and environments. In reality, the sources of variation include; differences in calibration, changes to setup and sensors~\cite{hosny2018artificial}, product changes, outliers from unforeseen events, etc.
This means that most systems are guarantied to encounter input that differs from the training distribution, likely resulting in faulty predictions~\cite{DBLP:journals/corr/abs-1811-11553}.

The idea presented here is analogous to a doctor looking at medical images in order to diagnose a patient. Here, similar images from previously confirmed cases are a big help in guiding a diagnosis. The idea of assisting doctors in this way has actually been implemented using an automatic image similarity search system~\cite{cai2019human}.
Similarly known samples are consulted when assigning a confidence to a new prediction. In the majority of related work looking at measuring image similarity, the problem often is that an algorithms understanding of "similar" is different from what is considered relevant by humans. What is interesting here is what is considered similar to the DNN. This is best measured using the learned representations from the activations in the layers of the DNN. By considering known samples that are similar in terms of activations across the DNN, we can get an idea of what performance to expect. The proximity to known samples can provide a level of certainty about the expected outcome. This idea rely on the assumption that the network is relatively smooth/stable locally.

Our research is preoccupied with automating processes found in slaughterhouses. The complexity of these processes and their non-stationary nature means that models must continuously adapt and improve or at least be able to predict when a given input is likely to result in a faulty prediction. The disruption and the monetary consequences of mistakes, makes it important for models to abstain and ask for guidance in cases where the outcome is unlikely to be acceptable. 
The purpose of this work is to detect out-of-distribution (OoD) samples for the tool pose predictor presented in~\cite{philipsen2019cutting}. This is used to judge whether predictions can be trusted and as a method for identifying the most valuable samples for extending the training set.

\subsection{Contributions}
With this paper, we address the problem of imperfect ML models in critical real world applications by determining whether a given prediction can be trusted. This is done based on the distance between the sample and nearby known samples in the feature space of a Deep Neural Network. 
The contributions can be summarized as:
\begin{enumerate}
    \item Confidence measure for unseen samples based on distance to training set neighbours in feature space.
\end{enumerate}

\section{Related Work}
Trust or confidence in predictions is important when choosing to deploy a new model and when predictions result in critical actions. Local Interpretable Model-agnostic Explanations (LIME), is a novel explanation technique that learn to provide an interpretable explanation of the predictions of any classifier. LIME enables a human to give feedback on the "reasoning" behind a prediction and suggest features to be removed from the model, leading to better generalization. The flexibility of the method is shown by applying it to random forests for text classification and neural networks for image classification. The idea is that global trust in the model is build by understand and gaining trust in individual predictions covering the input space. Although the complete model might be difficult to explain, individual decisions are explainable by relying on a local neighborhood region. The method is time consuming, taking up to 10 minutes to explain predictions for an image classification task~\cite{DBLP:journals/corr/RibeiroSG16}.
A small “loss prediction module” can be added to a target network, the module then learns to predict target losses of unlabeled inputs and thereby the ability to predict whether samples are likely to result in wrong predictions from the target model. The module is connected to several layers of the target model thereby considering different levels of information for loss prediction. The loss of the target model constitutes the ground truth for the loss prediction module~\cite{DBLP:journals/corr/abs-1905-03677}.
Using the data embedding from one of the final layers of the network a confidence score is computed based on local density estimates. Although, the method can be used for any DNN, it does require changes to the training procedure in order for the scores to be useful~\cite{DBLP:journals/corr/abs-1709-09844}.

Low confidence samples can be considered as being OoD. When such samples are detected, a system can either query for help~\cite{DBLP:journals/corr/abs-1902-10363} or abstain from predicting as done for detection of liver abnormalities using ultrasound images~\cite{DBLP:journals/corr/abs-1811-04463}.


\section{Method}
The presented method relies on the latent representation is found in the layers of DNNs. The latent representations correspond to the networks activations at a given layer and contain some of the information originally expressed in the input. The representations can be considered as coordinates in a high dimensional space. The further down the DNN the representation is found, the higher level concepts it describes. Representations can be extracted from any DNN and at any layer, but it is preferable to rely on a bottleneck or one of the final layers in order to lower the dimensionality of the latent space.

\begin{figure}[t]
  \centering
  \includegraphics[width=1.0\columnwidth]{\paperPath 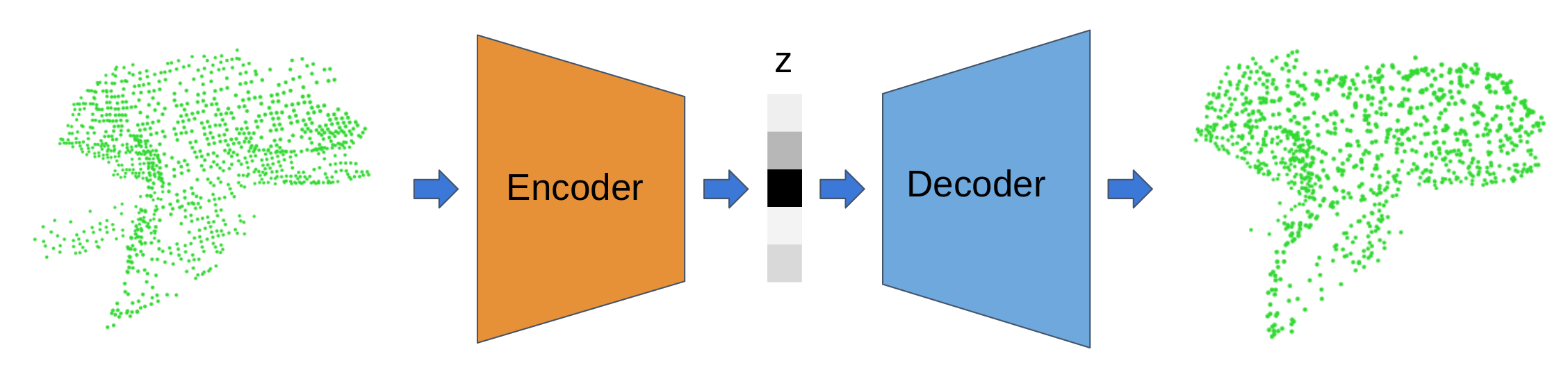}
  \caption{Autoencoder architecture with input point cloud, latent space bottleneck $z$, and reconstructed point cloud.}
  \label{method:point_cloud_ae}
\end{figure}

Here an autoencoder network architecture (see Figure \ref{method:point_cloud_ae}) is used to learn latent representations of point clouds. The autoencoder is based on PointNet~\cite{DBLP:journals/corr/QiSMG16} and has been used for recognition tasks and shape editing~\cite{DBLP:journals/corr/AchlioptasDMG17}. Principal component analysis is used to reduce the dimensionality of the representations, primarily for visualization purposes. Only the two most principal components are used in the following plots.

\subsection{Measure of Similarity}
The underpinning intuition for comparing samples using their latent representations is that similar inputs elicit similar activations throughout the DNN. 
Figure~\ref{method:density_n_confidence} (a) shows the distribution of known samples i.e. training samples (blue) and new unknown samples based on their latent representations. This reveals that the two data sets are very similar.

\subsection{Distribution of Errors}
\begin{figure}[h]
\centering
\begin{subfigure}{0.49\columnwidth}
  \centering
  \includegraphics[width=1.0\linewidth]{\paperPath 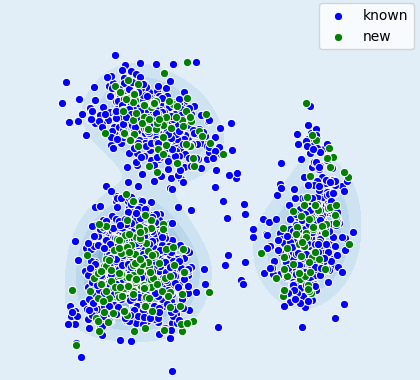}
  \caption{}
  \label{method:density_n_confidence:fig:density_rec}
\end{subfigure}
\hfill
\begin{subfigure}{0.49\columnwidth}
  \centering
  \includegraphics[width=1.0\linewidth]{\paperPath 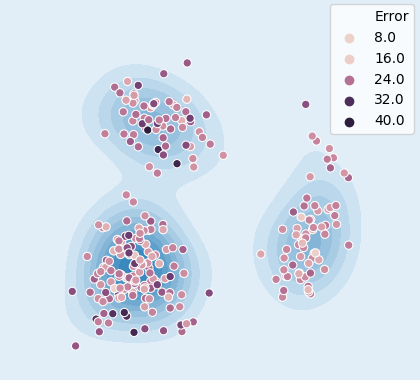}
  \caption{}
  \label{method:density_n_confidence:fig:confidence_rec}
\end{subfigure}
\caption{(a) Known samples (blue) and new samples (green) based on their feature space representations. (b) Distribution of error rates for new samples.}
\label{method:density_n_confidence}
\end{figure}

It is interesting to investigate whether samples with large errors are found in specific areas of the latent space. In this case the error is the reconstruction loss of the autoencoder, for other tasks and network types, the error can be any other kind of loss or prediction error that can be quantified. Figure~\ref{method:density_n_confidence} (b) shows the reconstruction error for new samples and their placement in the feature space.

\subsection{Error as a Function of Distance}
\begin{figure}[t]
  \centering
  \includegraphics[width=1.0\linewidth]{\paperPath 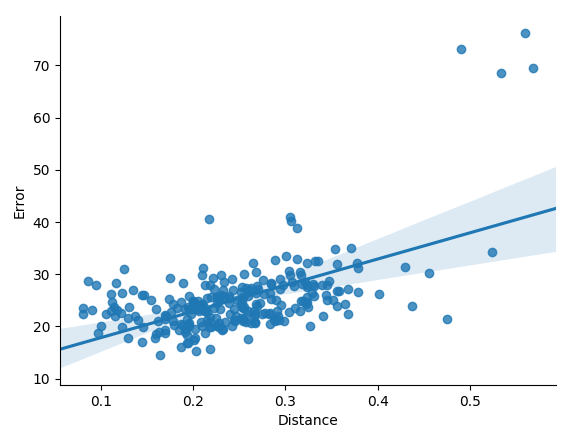}
  \caption{The reconstruction error in relation to the distance to nearest neighbor in the training set.}
  \label{method:dist_vs_err_rec}
\end{figure}

It is practically impossible to make guaranties about the performance of DNNs on unseen samples. This is especially the case with limited amounts of training data or OoD samples.
Figure~\ref{method:dist_vs_err_rec} shows the reconstruction error for new samples in relation to their distance from known samples. There is a clear trend showing a direct relationship between error and distance to nearest neighbor. This means that a threshold can be used to classify OoD samples as OoD.

\subsection{Identifying Out-of-distribution Samples}
\begin{figure}[h]
\centering
\begin{subfigure}{1.0\columnwidth}
  \centering
  \includegraphics[width=1.0\linewidth]{\paperPath 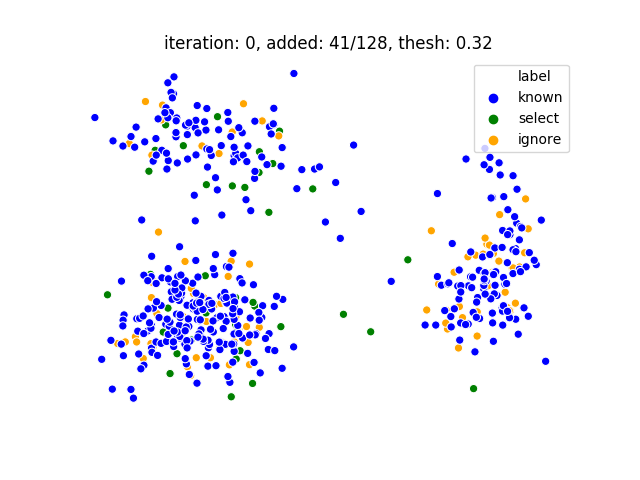}
  \caption{}
  \label{method:selection_from_confidence:fig:batch0}
\end{subfigure}
\newline
\begin{subfigure}{1.0\columnwidth}
  \centering
  \includegraphics[width=1.0\linewidth]{\paperPath 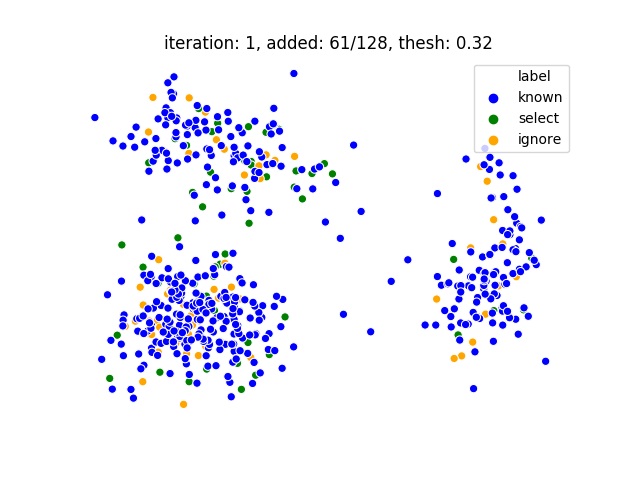}
  \caption{}
  \label{method:selection_from_confidence:fig:batch128}
\end{subfigure}
\caption{Training set distribution (blue), novel samples that should be added to training distribution (green), samples that contain insufficient novelty to warrant adding to training distribution (orange), novel samples that are outside of the confidence threshold and should be abstained from as well as added to the training distribution (red), (a) First batch. (b) Second batch.}
\label{method:selection_from_confidence}
\end{figure}

For a system that is being deployed in a critical production environment, the threshold may be based on the severity of error that can be accepted. In a less critical environment or during the introduction of a new system, the threshold may be determined by the amount of human involvement that can be afforded.
Figure~\ref{method:selection_from_confidence} (a) and (b) shows the gradual expansion of a training set based on a threshold that is selected based on a given labeling budget. 

\section{Discussion}
By measuring the similarity between an unseen sample and known samples it is possible to anticipate how the system will perform on the new sample. Knowing whether a given new sample exist in a sparse region of the training distribution and if performance in the region is acceptable may be useful when trying to determine whether to act on a prediction. This knowledge can also be used to select the most critical new examples to be added to the training set.
This is a tool which has the potential to enable cheaper, earlier, and safer deployment of models. It is vital when deploying models in ever-changing environments.






\section*{Acknowledgment}
This work was supported by Innovation Fund Denmark and the Danish Pig Levy Fund.

\bibliographystyle{IEEEtran}
\bibliography{main}

\end{document}